\documentclass[lettersize,journal]{IEEEtran}
\usepackage{amsmath,amsfonts}
\usepackage{array}
\usepackage[caption=false,font=normalsize,labelfont=sf,textfont=sf]{subfig}
\usepackage{textcomp}
\usepackage{stfloats}
\usepackage{url}
\usepackage{verbatim}
\usepackage{graphicx}
\usepackage{mathtools}
\usepackage{amssymb}
\usepackage{booktabs}
\usepackage{float}
\usepackage{algorithm}
\usepackage{algpseudocode}
\usepackage{multirow}
\usepackage[hidelinks]{hyperref}

\makeatletter
\renewcommand{\fs@ruled}{%
  \def\@fs@cfont{\bfseries}%
  \let\@fs@capt\floatc@ruled
  \def\@fs@pre{\kern5pt\hrule height.8pt depth0pt\kern2pt}%
  \def\@fs@post{\kern2pt\hrule\relax}%
  \def\@fs@mid{\kern2pt\hrule\kern2pt}%
  \let\@fs@iftopcapt\iftrue
}
\makeatother

\def\BibTeX{{\rm B\kern-.05em{\sc i\kern-.025em b}\kern-.08em
    T\kern-.1667em\lower.7ex\hbox{E}\kern-.125emX}}
\usepackage{balance}

\author{Michal Ciebielski$^{1}$, 
Shafeef Omar$^{1}$, Aaron Johnson$^{2,3}$, Majid Khadiv$^{1}$ % <-this % stops a space
\thanks{This work was funded in part by SIEMENS AG and the Technical University of Munich - Institute for Advanced Study and by the Huawei-TUM joint laboratory.}% <-this % stops a space
\thanks{$^{1}$Munich Institute of Robotics and Machine Intelligence (MIRMI), Technical University of Munich (TUM), Germany. {\tt\small firstname.lastname@tum.de}}
\thanks{$^{2}$Institute for Advanced Study, Technical University of Munich, Garching, Germany.}
\thanks{$^{3}$Department of Mechanical Engineering, Carnegie Mellon University, Pittsburgh, PA, USA. {\tt\small amj1@andrew.cmu.edu}}
}

\begin{document}
\title{FARO: Feasibility-Aware Robot Motion Optimization}
% \title{FARO: Feasibility-Guided Motion Generation for Loco-Manipulation}
% \title{Feasibility-Aware Motion Generation for Loco-Manipulation}

% \markboth{Journal of \LaTeX\ Class Files,~Vol.~18, No.~9, September~2020}%
% {How to Use the IEEEtran \LaTeX \ Templates}

% \IEEEaftertitletext{%
%   \vspace{-1ex}
%   \begin{center}
%     \includegraphics[width=0.7\textwidth]{figures/thumbnail.pdf}

%     %\smallskip
%     \refstepcounter{figure}\label{fig:teaser}%
%     {\footnotesize Fig.~\thefigure. \textbf{Real-world humanoid motion from FARO.}
%     Snapshots of a real-world experiment executing a trajectory found by our motion optimization.}
%   \end{center}
%   % \vspace{-1ex}
% }
\maketitle
\begin{abstract}
Fast planning of novel behaviors in unseen scenarios remains a fundamental challenge in robotics. The high-dimensional, hybrid, and underactuated nature of humanoid loco-manipulation continues to hinder the realization of this goal. In this paper, we address this challenge by proposing a nested kino-dynamic framework for rapid feasibility checking and dynamically consistent trajectory generation given a candidate contact sequence. By integrating this module with a feasibility-guided tree search and a Large Language Model (LLM)-based contact plan sampling strategy, we demonstrate that the proposed framework can substantially improve the search process. Furthermore, we show that the generated trajectories can be tracked using a reinforcement learning (RL)-based controller and show that the resulting trajectories are of sufficiently high quality for execution in real-world loco-manipulation scenarios. A supplementary video is available at: \url{https://youtu.be/R6qCHoCormQ}.
\end{abstract}

\begin{IEEEkeywords}
Multi-contact whole-body motion planning, constrained motion planning, kino-dynamic feasibility, fast motion discovery
\end{IEEEkeywords}

\section{Introduction}
\label{sec:introduction}
Reasoning about the environment and planning in novel situations are hallmarks of intelligence. Evidence suggests that apes \cite{vaesen2012cognitive} and crows \cite{wimpenny2009cognitive} are capable of such planning behaviors, particularly in tool-use tasks. Replicating these capabilities in robotics has proven extremely challenging, primarily because it requires exploration over both motion and feasible force spaces, resulting in a highly complex planning problem. The challenge becomes even more pronounced in humanoid loco-manipulation, where the problem additionally involves searching through an extremely high-dimensional space of the underactuated system over a long horizon.

Trajectory optimization (TO) and reinforcement learning (RL) are the two primary paradigms for realizing such behaviors, and both have seen significant progress over the past decade \cite{wensing2024optimization,ha2025learning}. In trajectory optimization, to avoid issues associated with overly smoothing the hybrid nature of contact interactions \cite{tassa2012synthesis, mordatch2012discovery}, most approaches combine discrete search over contact sequences with trajectory optimization \cite{tonneau2018efficient,toussaint2018differentiable,sleiman2023versatile,ciebielski2025task}, which can be viewed as a relaxation of mixed-integer formulations \cite{deits2014footstep,aceituno2017simultaneous,ponton2021efficient}. Although these methods improve contact reasoning and behavioral diversity, the combinatorial growth of the search space makes them computationally expensive. Maintaining tractability therefore often requires restricting the search through simplified models, task-specific heuristics, or hand-designed high-level actions, limiting the range of complex dynamic behaviors that can be discovered.

On the other hand, RL has achieved remarkable success in locomotion~\cite{ha2025learning} due to the cyclic nature of the task and thanks to the extreme human input through extensive reward shaping \cite{hwangbo2019learning,lee2020learning}, but it remains ineffective for solving complex long-horizon loco-manipulation skills. Consequently, many recent advances in RL-based humanoid loco-manipulation rely on tracking demonstrations~\cite{seo2023deep} or human-to-humanoid retargeting \cite{pan2025spiderscalablephysicsinformeddexterous,yang2025omniretargetinteractionpreservingdatageneration,dhedin2026dynaretargetdynamicallyfeasibleretargetingusing} to bypass the difficult exploration problem. While effective when suitable reference motions are available, these approaches bypass the problem of discovering behaviors in novel situations.

Exploration thus remains the central obstacle to the generation of novel loco-manipulation behaviors: its scope is restricted in TO to manage computational complexity and largely avoided in RL through imitation or retargeting. Our approach instead addresses this challenge through more efficient exploration of multi-contact motion. By rapidly pruning infeasible contact decisions, it reduces the combinatorial burden of search without relying on predefined behavioral abstractions or reference motions.
\begin{figure}
    \centering
    \includegraphics[width=0.475\textwidth]{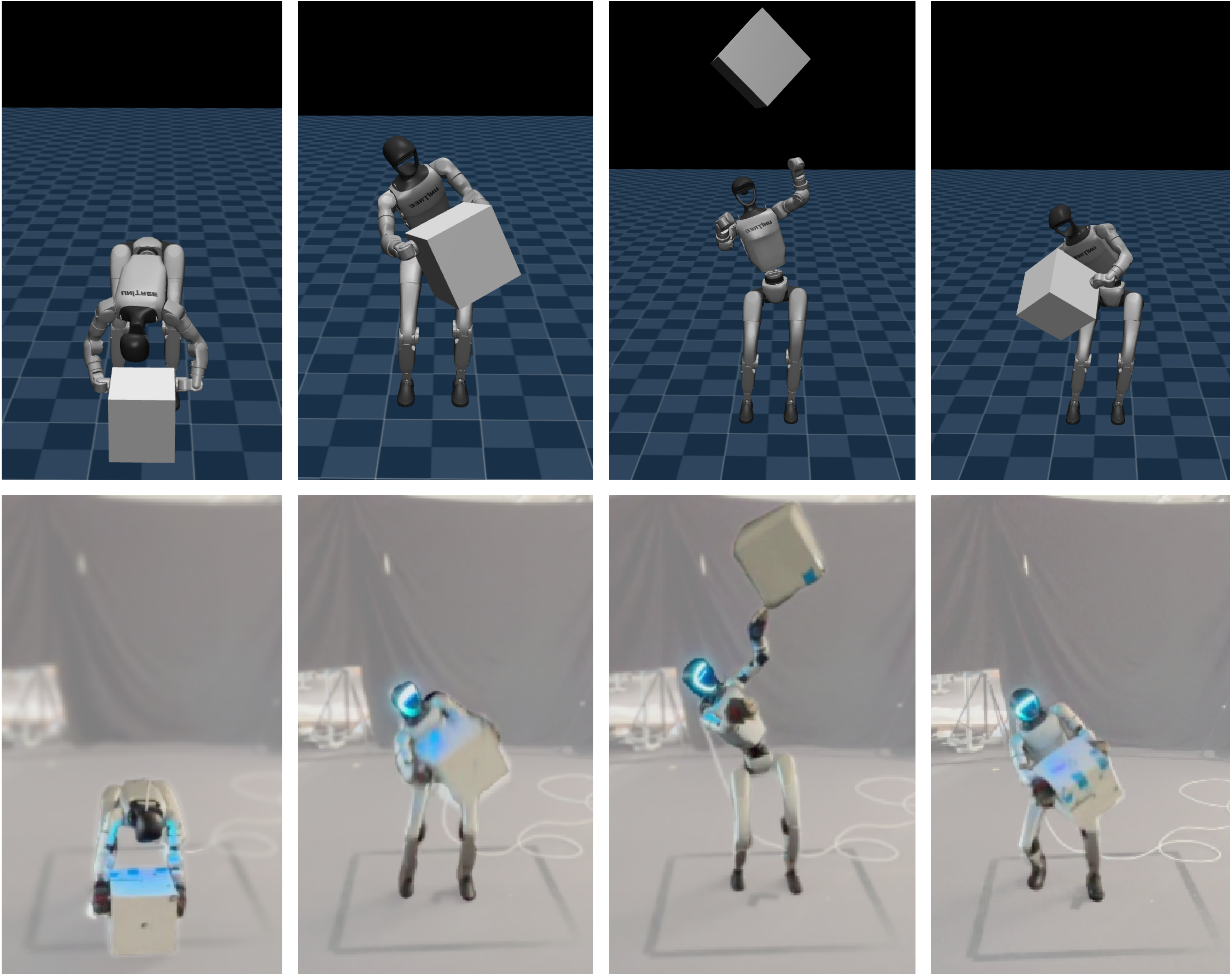}
    % \vspace{-4mm}
    \caption{\textbf{Motion optimization and real-world execution with \textit{FARO}.} Top: snapshots of a trajectory produced by our motion optimization. Bottom: snapshots from executing the optimized trajectory on the real robot.}
    \label{fig:teaser}
\end{figure}

\begin{figure*}
    \vspace*{2pt}
    \centering
    \includegraphics[width=1.0\textwidth]{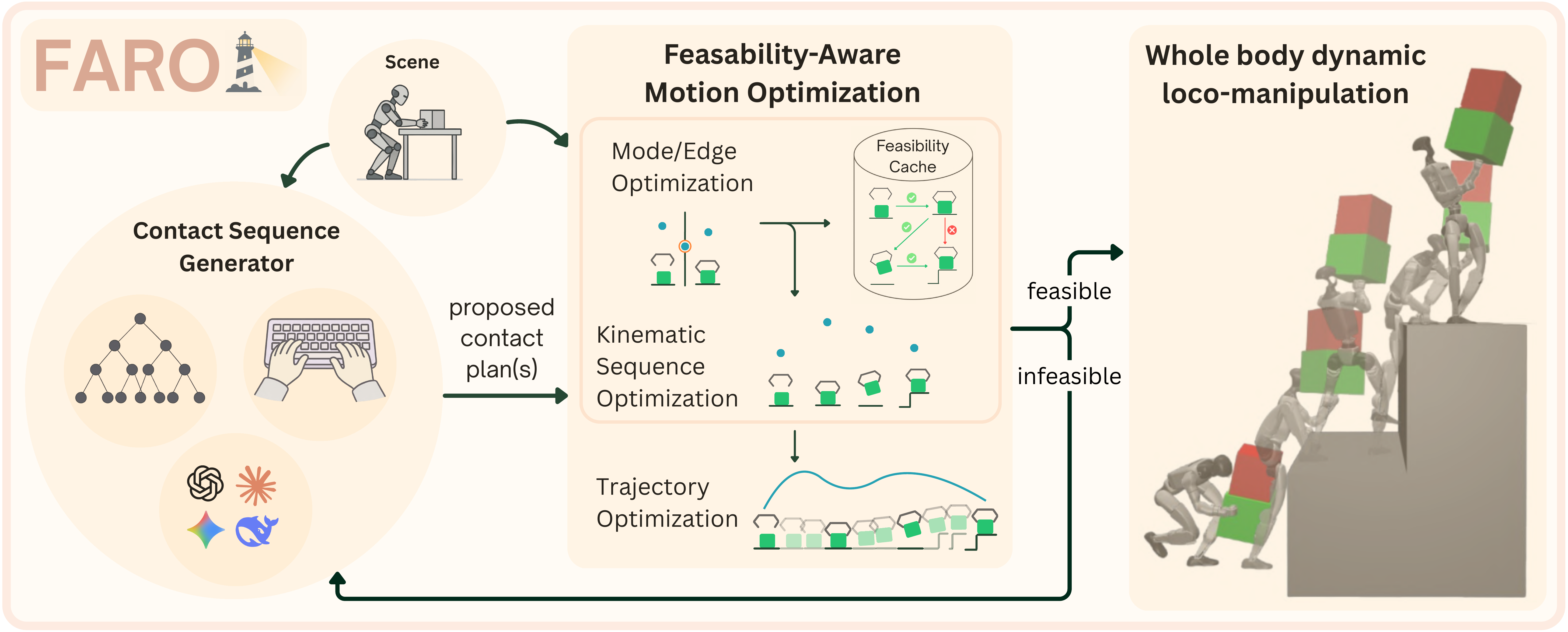}
    % \vspace{-4mm}
    \caption{\textbf{High-level overview of \textit{FARO}.} The scene defines the robot, movable objects, environment geometry, and available contact interfaces from which candidate contact-mode sequences are generated. The contact sequence generator may use search-based, user-specified, or LLM-based methods to produce candidate contact plans. These plans are evaluated by a hierarchy of increasingly restrictive feasibility checks consisting of mode/edge optimization, kinematic sequence optimization, and trajectory optimization. Results of mode/edge feasibility checks are stored in a cache, allowing previously evaluated modes and transitions to be accepted or rejected without resolving the corresponding optimization. Dynamically feasible plans produce whole-body, multi-object loco-manipulation motions, while infeasible plans can provide feedback to the contact sequence generator, which continues proposing and evaluating new contact sequences.}
    \label{fig:framework_overview}
\end{figure*}

In this paper, we propose a general exploration strategy for efficiently searching the space of motion and contact interaction in humanoid loco-manipulation. Our framework, Feasibility-Aware Robot Motion Optimization \textit{(FARO)}, summarized in Fig.~\ref{fig:framework_overview}, combines progressively more expressive optimization-based feasibility tests with search to explore discrete contact-mode sequences while preserving a broad range of dynamically feasible behaviors. We demonstrate the framework in model-based tree search, LLM-guided contact-plan generation, and plans specified through human input. The resulting dynamic loco-manipulation skills are transferred to a real humanoid robot using a pre-trained RL-based trajectory-tracking controller, as illustrated in Fig.~\ref{fig:teaser}. This work represents a step toward the efficient, on-the-fly generation of complex loco-manipulation behaviors in real-world environments.

The contributions of this paper are:
\begin{itemize}
\item a novel optimization-based pruning strategy, \textit{FARO}, for multi-contact motion planning that substantially accelerates search without introducing a significant number of false-negative feasibility classifications;

\item the application of the \textit{FARO} pruning strategy to a feasibility-guided tree search algorithm for multi-contact motion planning that substantially outperforms direct trajectory-optimization-based exploration;

\item a systematic evaluation of the \textit{FARO} approach in three different algorithms: the model-based tree search, an LLM-guided contact-plan generation, and contact plans provided through human input.

\end{itemize}

% \section{Related Work}
% \label{sec:related_work}

%===============================================================================

\section{Contact-Explicit Optimization}
\label{sec:method}
\begin{figure*}
    \vspace*{2pt}
    \centering
    \includegraphics[width=0.9\textwidth]{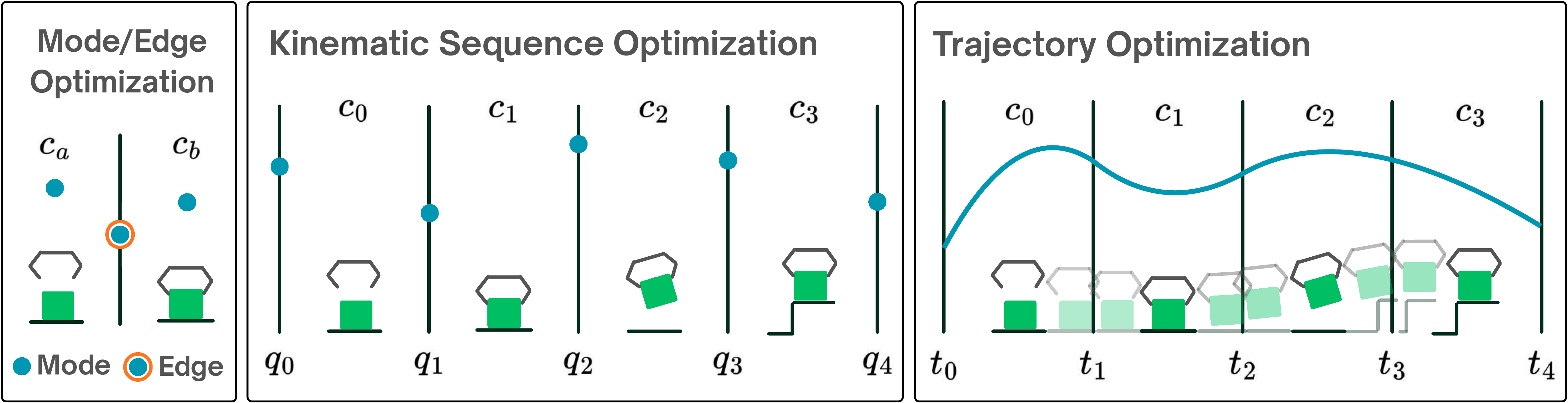}
    % \vspace{-4mm}
    \caption{\textbf{Graphical overview of the optimization problems used in \textit{FARO}}. The mode/edge optimization finds a single configuration that satisfies the constraints associated with an individual contact mode or transition. The kinematic sequence optimization finds one configuration per transition plus one configuration satisfying the initial and one the terminal condition. The trajectory optimization searches for a dynamically feasible motion by discretizing a continuous trajectory over the contact-mode sequence and enforcing the scene dynamics, contact constraints, and initial and terminal conditions throughout the trajectory. These checks form a hierarchy of increasingly restrictive feasibility tests: a feasible contact-mode sequence requires its constituent modes and transitions to be feasible, while dynamic feasibility further requires the existence of a feasible kinematic sequence. Thus, moving from left to right, feasibility at one level is necessary, but not sufficient, for feasibility at the next. Because the necessary conditions on the left are substantially cheaper to evaluate than full trajectory optimization, they can reject infeasible candidates early and avoid unnecessary trajectory optimizations, thereby accelerating  search, as demonstrated in our evaluation.}
    \label{fig:step_comparison}
\end{figure*}

Our method centers on contact-explicit optimization, in which a discrete contact-mode sequence is provided as input and the corresponding continuous state, control, and timing variables are optimized subject to the prescribed contacts. To efficiently identify promising contact sequences, we complement a full trajectory optimization with progressively stronger feasibility checks that operate on individual modes, transitions, and complete kinematic sequences. These optimization problems, visualized graphically in Fig~\ref{fig:step_comparison}, share a common set of contact, collision, dynamics, and limit constraints, while differing in the variables they optimize and the level of feasibility they enforce.

We first define the contact-mode sequence representation in Section~\ref{subsec:contact_mode} and the state and control spaces in Section~\ref{sec:state_control_space}, followed by the common constraint formulations in Section~\ref{sec:constraint_definitions}. The mode/edge feasibility optimization, kinematic sequence optimization, and full trajectory optimization are then presented in Sections~\ref{sec:pruning_optimization}, \ref{sec:seqik_optimization}, and \ref{sec:trajectory_optimization}, respectively. Following this Section~\ref{sec:search} describes how these optimization-based feasibility checks are integrated into a feasibility-guided contact-mode tree search.
\subsection{Contact Mode Sequence Definition}
\label{subsec:contact_mode}

Let \(\mathcal{I}\) be the set of all contact interfaces in the scene, indexed by integers.
%, and let \(\mathcal{I}^{\mathrm{act}} \subseteq \mathcal{I}\) denote the active interfaces whose contact state may change. These include, for example, end-effector interfaces as well as interfaces attached to movable and static objects.
For each interface \(a \in \mathcal{I}\), we represent its contact state by a pair \((a,b)\), where \(b \in \mathcal{I}\) indicates unilateral contact with interface \(b\), and \(b=\varnothing\) indicates that \(a\) is free. A contact mode is the complete assignment over all interfaces a unique contact pair:
\begin{equation}
c =
\{(a,b) : \forall a \in \mathcal{I}\ \exists! \;
b \in \mathcal{I} \cup \{\varnothing\}\}.
\end{equation}
Furthermore, the mode must be consistent, i.e.\ $(a,b) \in c \Leftrightarrow (b,a) \in c$.
For a contact-mode time sequence \(\mathcal{C} = (c_0,\ldots,c_{K-1})\), each mode at time $s$ is written as
\begin{equation}
c_s =
\{(a,b_{s}) : a \in \mathcal{I}\},
\qquad s = 0,\ldots,K-1 .
\end{equation}
\subsection{State and Control Spaces}
\label{sec:state_control_space}

The state space and control variables are defined as
\begin{align}\label{eq:system_variables}
%\begin{gather}
%q := (q^r,q^{o_1},\ldots,q^{o_{n_o}}) \in\mathcal Q,
%\label{eq:system_configuration}
%\\
x
:=
(x^r,x^{o_1},\ldots,x^{o_{n_o}})
\in\mathcal X,
\\
u
:=
(u^r,u^{o_1},\ldots,u^{o_{n_o}})
\in\mathcal U .
%\label{eq:system_state_control}
%\end{gather}
\end{align}
with components corresponding to the robot, $r$, and objects, $o_\ell$, as indicated by the superscript,
\begin{subequations}\label{eq:state_control_components}
\begin{alignat}{2}
x^r
&:=
(q^r,v^r,h^r),
\quad&
u^r
&:=
(\dot v^r,\lambda_{1}^r,\ldots,\lambda_{n_{ee}}^r),
\label{eq:state_components}
\\
x^{o_\ell}
&:=
(q^{o_\ell},\mathcal V^{o_\ell}),
\quad&
u^{o_\ell}
&:=
(\dot{\mathcal V}^{o_\ell},\mathcal W_{\mathrm{env}}^{o_\ell}).
\label{eq:control_components}
\end{alignat}
\end{subequations}
where \(q^r\), \(v^r\), and \(h^r\) denote the robot configuration, generalized velocity, and centroidal momentum, respectively. For each object \(o_\ell\), \(q^{o_\ell}\) and \(\mathcal V^{o_\ell}\) denote its pose and body velocity. The robot control consists of the generalized acceleration \(\dot v^r\) and end-effector wrenches \(\lambda_{e}^r\), \(e=1,\ldots,n_{ee}\), while each object control consists of the body acceleration \raisebox{-0.20ex}{\scalebox{0.90}{\ensuremath{\dot{\mathcal V}^{o_\ell}}}} and external environment wrench \(\mathcal W_{\mathrm{env}}^{o_\ell}\).
The component spaces are
\begin{subequations}\label{eq:component_spaces}
\begin{gather}
q^r \in \mathrm{SE}(3)\times\mathbb R^n,\quad
v^r \in \mathfrak{se}(3)\times\mathbb R^n,\quad
h^r \in \mathbb R^6,
\notag
\\
\dot v^r \in \mathfrak{se}(3)\times\mathbb R^n,\quad
\lambda_{e}^r \in \mathbb R^6,
\label{eq:robot_component_spaces}
\\
q^{o_\ell} \in \mathrm{SE}(3),\
\mathcal V^{o_\ell} \in \mathfrak{se}(3),\
\dot{\mathcal V}^{o_\ell} \in \mathfrak{se}(3),\
\mathcal W_{\mathrm{env}}^{o_\ell} \in \mathbb R^6 .
\label{eq:object_component_spaces}
\end{gather}
\end{subequations}
\subsection{Constraint Definitions}
\label{sec:constraint_definitions}
\subsubsection{Contact}
In this work, all end-effectors and environment contact interfaces are modeled as rectangular patches with patch-to-patch unilateral contact, though other contact modalities, such as sliding or bilateral contact, could be incorporated. This includes contact between end-effector patches, patches on movable objects, and static environment patches. 

Let \(p\), \(R\), \(f\), and \(\kappa\) denote the relative position, rotation, force, and moment of contact patch \(a\) with respect to contact patch \(b\), expressed in frame \(b\). The patch normals are aligned with their local \(z\)-axes. The vector \(\xi\) denotes the half-extents of a rectangular patch, so \(\prescript{b}{}{\xi}_{a}\) denotes the half-extents of patch \(a\) expressed in frame \(b\). The instantaneous constraints for patch-patch contact are:
\begin{subequations}\label{eq:patch_to_patch_constr}
\begin{gather}
\log_3\left(
    R
\right)_{x,y}
= 0,
\quad
p_z
= 0,
\label{eq:patch_patch_planar}
\\
\left|
    p_{x,y}
\right|
\leq
\left(
    \prescript{b}{}{\xi}_{b}
\right)_{x,y}
-
\left(
    \prescript{b}{}{\xi}_{a}
\right)_{x,y},
\label{eq:patch_patch_bounds}
\\
\left|
    f_{x,y}
\right|
\leq
\mu
f_z,
\quad
f_z
\geq 0,
\label{eq:contact_force_constraints}
\\
\left|
    \kappa_z
\right|
\leq
\mu_r
f_z,
\quad
\left|
\begin{bmatrix}
    \kappa_x \\
    \kappa_y
\end{bmatrix}
\right|
\leq
f_z
\begin{bmatrix}
    \left(\prescript{a}{}{\xi}_{a}\right)_y \\
    \left(\prescript{a}{}{\xi}_{a}\right)_x
\end{bmatrix}.
\label{eq:contact_moment_constraints}
\end{gather}
\end{subequations}
Constraint~\eqref{eq:patch_patch_planar} aligns the patch normals through \(\log_3(R)\) and enforces zero normal separation, while leaving the relative in-plane position and yaw angle free. Constraint~\eqref{eq:patch_patch_bounds} keeps patches within the admissible region defined by the half-extents \(\xi\). Constraint~\eqref{eq:contact_force_constraints} enforces unilateral contact and a pyramidal approximation of Coulomb friction with coefficient \(\mu\). Constraint~\eqref{eq:contact_moment_constraints} enforces torsional friction with coefficient \(\mu_r\) and center-of-pressure bounds induced by the finite patch size.
For a sticking contact that remains active across adjacent timesteps \(s\) and \(s+1\), we also have the no-slip condition,
\begin{equation}\label{eq:sticking_position}
\begin{aligned}
\left(
    p^{s+1}
    -
    p^{s}
\right)_{x,y}
&= 0,
\quad
\log_3\left(
    \left(R^{s}\right)^\top
    R^{s+1}
\right)_z
= 0.
\end{aligned}
\end{equation}
which is enforced by constraining both the in-plane position and the in-plane rotation across timesteps.

\subsubsection{Collision} All bodies in the robot and scene are modeled using convex geometries, and collision avoidance is imposed with the signed-distance approximation of Schulman et al.~\cite{schulman2014motion}. For each collision pair \(A\) and \(B\), the collision is queried using the GJK algorithm~\cite{gilbert1988fast} as implemented in \cite{coalweb}. The query returns witness points \(p_A\) and \( p_B\), expressed in the local frames of the two bodies, together with a separating normal \(\hat{ n}\). $T_A^w(x)$ and $T_B^w(x)$ represent the transformation of body $A$ and $B$ in the world frame. The resulting constraint based on the  signed-distance approximation is
\begin{equation}\label{eq:collision_constr}
    0 \leq\operatorname{sd}_{AB}(x)
    \approx
    \hat{ n}
    \cdot
    \left(
    T_A^w(x) p_A
    -
    T_B^w(x)  p_B
    \right)
\end{equation}

\subsubsection{Dynamics}
The robot dynamics constraints are,
\begin{subequations}\label{eq:robot_dynamics}
\begin{equation}\label{eq:robot_dynamics_qvh}
\begin{gathered}
    q^r_{i+1}
    =
    q^r_{i} \oplus v^r_{i+1}\,(\bar T\Delta t), \quad
    v^r_{i+1}
    =
    v^r_{i} + \dot{v}^r_{i+1}\,(\bar T\Delta t),
    \\
    h^r_{i+1}
    =
    h^r_i + \dot{h}^r_{i+1}\,(\bar T\Delta t),
\end{gathered}
\end{equation}
\begin{equation}\label{eq:robot_dynamics_hdot}
\begin{gathered}
    \dot{h}^r
    =
    \begin{bmatrix}
        mg + \sum_{e=1}^{n_{ee}} f_{e}^r \\
        \sum_{e=1}^{n_{ee}} (p_{e}^r - c) \times f_{e}^r + \kappa_{e}^r
    \end{bmatrix},
    \quad
    h^r
    =
    A(q^r)v^r.
\end{gathered}
\end{equation}
\end{subequations}
where \eqref{eq:robot_dynamics_qvh} integrates the robot configuration, velocity, and centroidal momentum using a backward Euler discretization over the scaled timestep \(\bar T \Delta t\), and \(\oplus\) denotes integration on the configuration manifold. The centroidal momentum rate \(\dot h^r\) in \eqref{eq:robot_dynamics_hdot} is given by the net external wrench acting on the robot, including gravity and the end-effector wrench components \(\lambda_{e}^r=(f_{e}^r,\kappa_{e}^r)\). Finally, the centroidal momentum \(h^r\) is constrained to ensure consistency with the whole-body robot dynamics via the centroidal momentum matrix $A(q^r)$.

The object dynamics are defined as,
\begin{subequations}\label{eq:object_dynamics}
\begin{equation}\label{eq:object_dynamics_qv}
\begin{gathered}
    q^{o_\ell}_{i+1}
    =
    q^{o_\ell}_i \oplus \mathcal{V}^{o_\ell}_{i+1}\,(\bar T\Delta t),
    \
    \mathcal{V}^{o_\ell}_{i+1}
    =
    \mathcal{V}^{o_\ell}_i + \dot{\mathcal{V}}^{o_\ell}_{i+1}\,(\bar T\Delta t)
\end{gathered}
\end{equation}
\begin{equation}\label{eq:object_dynamics_wrench}
\begin{gathered}
    \mathcal{W}^{o_\ell}_{\mathrm{ext}}
    =
    \mathcal{G}^{o_\ell}\,\dot{\mathcal{V}}^{o_\ell}
    -
    \bigl[\mathrm{ad}_{\mathcal{V}^{o_\ell}}\bigr]^\top
    \mathcal{G}^{o_\ell}\,\mathcal{V}^{o_\ell}.
\end{gathered}
\end{equation}
\end{subequations}
where \eqref{eq:object_dynamics_qv} integrates the object pose and twist using a backward Euler discretization over the scaled timestep \(\bar T \Delta t\), and \(\oplus\) denotes integration on the object configuration manifold. \eqref{eq:object_dynamics_wrench} defines the object twist--wrench dynamics driven by the total external wrench, following \cite{lynch2017modern}. Here, \(\mathcal{W}^{o_\ell}\) denotes wrenches expressed in the object body frame, \(\bigl[\mathrm{ad}_{\mathcal{V}^{o_\ell}}\bigr]\) is the adjoint Lie bracket operator associated with the object body twist \(\mathcal{V}^{o_\ell}\), and \(\mathcal{G}^{o_\ell}\) is the spatial inertia matrix of object \(o_\ell\).
$\mathcal{W}^{o_\ell}_{\mathrm{ext}}
:=
\mathcal{W}^{o_\ell}_{\mathrm{env}}
+
\mathcal{W}^{o_\ell}_{\mathrm{grav}}
+
\sum_{a\in\mathcal I}
\mathcal{W}^{o_\ell}_{a}$ collects the signed body wrenches entering the object dynamics: \(\mathcal{W}^{o_\ell}_{\mathrm{env}}\) encodes interaction with the environment, \(\mathcal{W}^{o_\ell}_{\mathrm{grav}}\) represents gravity, and \(\mathcal{W}^{o_\ell}_{a}\) captures the wrench applied through interface \(a\)  on object \(o_\ell\).

\subsubsection{Limits}
Let the actuated joint torque be
\begin{equation}
\tau_j := S \left( M(q)\dot v + b(q,v) - \sum_{e=1}^{n_{ee}}
J_{e}(q)^\top \lambda_{e}^r \right).
\label{eq:actuated_torque}
\end{equation}
where $S$ selects the actuated components of the dynamics. Similarly, $q_j$ and $v_j$ denote the vectors of actuated joint positions and velocities. The joint position, velocity, and torque-speed limits are
\begin{subequations}\label{eq:limit_constraints}
\begin{equation}\label{eq:position_limits}
q_{\min}
\leq q_j \leq q_{\max},
\end{equation}
\begin{equation}\label{eq:velocity_limits}
v_{\min}
\leq v_j \leq v_{\max}.
\end{equation}
\begin{equation}\label{eq:torque_speed_limits}
|\tau_j|
+
\frac{\tau_{\max}}{v_{\tau,\max}} |v_j|
\leq
\tau_{\max}.
\end{equation}
\end{subequations}
The last inequality is a linear torque-speed envelope: the available torque decreases affinely with joint speed, reaching zero at $v_{\tau,\max}$. Each constraint in \eqref{eq:limit_constraints} is applied componentwise.

\subsection{Contact Mode and Edge Feasibility}
\label{sec:pruning_optimization}
In order to prune kinematically infeasible contact modes and transitions between them, we formulate the following inverse kinematics optimization problem \eqref{eq:graph_pruning}. A contact mode $c$ is said to be infeasible if the resulting nonlinear program does not converge within a number of maximum iterations. To test whether two modes $c_1$ and $c_2$ can be connected via an edge, $c$ is replaced by $c_1 \cup c_2$, representing the instantaneous transition between the two modes.
\begin{equation}\label{eq:graph_pruning}
\begin{aligned}
\min_{q}
& \quad (q - q_{\mathrm{nom}})^\top W (q - q_{\mathrm{nom}})\\
\text{s.t.}\quad
& \mathrm{contact}(q,c) \leq 0 \quad \eqref{eq:patch_patch_planar},\eqref{eq:patch_patch_bounds},\\
& \mathrm{collision}(q) \leq 0 \quad \eqref{eq:collision_constr},\\
& \mathrm{limits}(q) \leq 0 \quad \eqref{eq:position_limits},\\
\end{aligned}
\end{equation}
The contact, collision, and limit constraints are defined in section \ref{sec:constraint_definitions}, while $q=(q^r,q^{o_1},\dots,q^{o_{n_o}})$ represents the robot and object configurations and $q_\mathrm{nom}$ represents the nominal configuration used for regularization with weight $W$. Within a search procedure, expansions from a parent node $c_\mathrm{parent}$ to candidate child nodes $c_\mathrm{child}$ are evaluated using this optimization, and the results are cached to avoid recomputing previously checked contact modes and transitions.

\subsection{Kinematic Sequence Optimization}
\label{sec:seqik_optimization}
Given a sequence of $K$ contact modes,
$\mathcal C=(c_0,\ldots,c_{K-1})$, the kinematic sequence optimization (KSO) solves for $K+1$ robot--object configurations $q_{0}=(q_0,\ldots,q_K)$. The configuration $q_0$ satisfies the initial condition, while $q_K$ can optionally constrain the terminal condition when evaluating a goal-reaching sequence. Each intermediate configuration $q_s$ represents the transition between adjacent modes $c_{s-1}$ and $c_s$. The optimization enforces the contact constraints \eqref{eq:patch_patch_planar}, \eqref{eq:patch_patch_bounds}, \eqref{eq:sticking_position}, collision avoidance \eqref{eq:collision_constr}, and configuration limits \eqref{eq:position_limits} along the sequence. For notational convenience, we define $c_K:=c_{K-1}$, so that the terminal configuration can be handled using the same indexing as the intermediate configurations. The resulting optimization is

\begin{equation}\label{eq:seqik} 
\begin{aligned} 
\min_{q_{0:K}} & \quad \sum_{s=0}^{K} (q_s-q_{\mathrm{nom}})^\top W (q_s-q_{\mathrm{nom}}) \\ \text{s.t.}\quad & q_0=q_{\mathrm{init}}, \quad q_K\in Q_{\mathrm{goal}}, \\
& \forall s\in\{1,\dots,K\}: \\ 
& \quad \mathrm{contact} \bigl(q_s,c_{s-1}\cup c_s\bigr) \leq 0 \quad \eqref{eq:patch_patch_planar}, \eqref{eq:patch_patch_bounds}, \\ 
& \quad \mathrm{collision}(q_s) \leq 0 \quad \eqref{eq:collision_constr}, \\ 
& \quad \mathrm{limits}(q_s) \leq 0 \quad \eqref{eq:position_limits}, \\ 
& \forall s\in\{0,\dots,K-1\}: \\ 
& \quad \forall a\in\mathcal I \text{ satisfying \eqref{eq:maintain_contact} or \eqref{eq:release_contact}}: \\ 
& \qquad \mathrm{contact} \bigl(q_s,q_{s+1},c_s) =0 \quad \eqref{eq:sticking_position}. 
\end{aligned} \end{equation}

The objective regularizes each configuration about the nominal configuration $q_\mathrm{nom}$ with weight $W$. The optimization tests geometric feasibility without imposing time-dependent dynamics. Coupling between adjacent configurations is introduced through transition-consistency constraints associated with sticking contacts. Specifically, for each transition $s \to s+1$, we enforce for any interface $a \in \mathcal{I}$ whose contact-state assignments in $c_s$ and $c_{s+1}$ satisfy either
\begin{subequations}\label{eq:transition_conditions}
\begin{align}
b_s &= b_{s+1}
\neq \varnothing,
\label{eq:maintain_contact}
\\
b_s &\neq \varnothing
\quad\land\quad b_{s+1}
= \varnothing.
\label{eq:release_contact}
\end{align}
\end{subequations}
Condition~\eqref{eq:maintain_contact} corresponds to persistent contact, while condition~\eqref{eq:release_contact} corresponds to contact release. In both cases, the relative in-plane pose of the contact interface is constrained to remain fixed across the transition according to~\eqref{eq:sticking_position}. For persistent contact, this preserves the contact location between adjacent configurations. For release, $q_{s+1}$ represents the boundary configuration at which contact is broken, so the sticking constraint is enforced up to the instant of separation.
\subsection{Trajectory Optimization}
\label{sec:trajectory_optimization}
Whereas the KSO verifies the geometric consistency of transitions along a given contact-mode sequence, the trajectory optimization additionally enforces the full robot--object dynamics over a discretized trajectory. Given the contact-mode sequence \(\mathcal C=(c_0,\ldots,c_{K-1})\), the optimization solves for the state trajectory \(x_{0:N}\), control trajectory \(u_{0:N-1}\), and one dimensionless time-scaling factor \(\bar T_s\) for each contact-mode stage \(s\):
\begin{equation}\label{eq:TO}
\begin{aligned}
\min_{\substack{x_{0:N},\,\\u_{0:N-1},\\\bar T_{0:K-1}}}
& \quad \sum_{i=0}^{N-1} \phi(x_i,u_i)
      + \phi_N(x_N)
      + w_T \sum_{s=0}^{K-1} (\bar T_s - 1)\\
\text{s.t.}\quad
& x_0 = x_{\mathrm{init}}, \ \ x_N \in X_{\mathrm{goal}},\\
& \forall s \in \{0,\dots,K-1\}:\\
& \quad \bar T_{\min} \leq \bar T_s \leq \bar T_{\max},\\
& \forall i \in \{0,\dots,N-1\}:\\
& \quad \mathrm{dyn}(x_i,x_{i+1},u_i,\bar T_{s(i)},c_{s(i)}) = 0 \quad \eqref{eq:robot_dynamics}, \eqref{eq:object_dynamics},\\
% & \quad \mathrm{contact}(x_i,u_i,c_{s(i)}) \in \mathcal{K}(c_{s(i)}),\\
& \quad \mathrm{contact}(x_i,u_i,c_{s(i)}) \leq 0 \quad \eqref{eq:patch_to_patch_constr}, \eqref{eq:sticking_position},\\
& \quad \mathrm{collision}(x_i) \leq 0 \quad \eqref{eq:collision_constr},\\
& \quad \mathrm{limits}(x_i,u_i) \leq 0 \quad \eqref{eq:limit_constraints}.
\end{aligned}
\end{equation}
Here, \(x_{\mathrm{init}}\) fixes the initial robot--object state, and \(X_{\mathrm{goal}}\) is the admissible terminal set for the task. The running and terminal costs are denoted by \(\phi\) and \(\phi_N\). The factor \(\bar T_s\) scales the nominal timestep \(\Delta t\) for all knot points in contact stage \(s\), so knot point \(i\) uses timestep \(\bar T_{s(i)}\Delta t\), where \(s(i)\) maps knot points to contact-mode stages. The bounds \(\bar T_{\min}\) and \(\bar T_{\max}\) constrain each stage duration, and \(w_T\sum_{s=0}^{K-1}(\bar T_s-1)\) penalizes deviations from the nominal scaling \(\bar T_s=1\). The remaining constraints enforce dynamics, contact feasibility, collision avoidance, and state/control limits at each knot point and are each described in section \ref{sec:constraint_definitions}.

\subsection{Implementation}
Both \eqref{eq:seqik} and \eqref{eq:TO} are implemented using a direct multiple-shooting transcription. The distance function approximation in \eqref{eq:collision_constr} is implemented with the collision and distance resolution in \texttt{coal} \cite{coalweb}. Robot kinematics and dynamics are evaluated with \texttt{Pinocchio} \cite{carpentier2019pinocchio}, and the resulting nonlinear programs are formulated in \texttt{CasADi} \cite{andersson2019casadi}. The KSO is solved with the SQP solver in \texttt{acados} \cite{Verschueren2021acados}, whereas the TO is solved with the SQP solver \texttt{Hippo} \cite{zhao2026hippo}. The mode/edge optimizations \eqref{eq:graph_pruning} are solved with \texttt{Ipopt} \cite{ipopt}.

\section{Feasibility-Guided Tree Search}
\label{sec:search}

Based on the optimization formulations presented in the preceding sections, we propose a feasibility-guided tree search over discrete contact-mode sequences, shown in Algorithm~\ref{alg:feasible_tree_search}. The tree is rooted at an initial mode $c_0$, and each node $v$, together with its root-to-node prefix, represents a contact-mode sequence $\mathcal C(v)$. Node selection uses a cost-based variant of Upper Confidence bounds applied to Trees (UCT), which balances the exploitation of low-cost branches with the exploration of less frequently visited nodes.
\begin{equation}
v^*
=
\operatorname*{argmin}_{u \in \mathrm{children}(v)}
\left[
J(u)
-
C
\sqrt{
\frac{\log(N(v)+1)}
{N(u)+1}
}
\,\right].
\end{equation}
$J(u)$ is the node cost, $C$ is the exploration coefficient and $N(\cdot)$ denotes the visit count. Expansion is controlled by progressive widening, with the maximum number of children $M(v)$ increasing as node $v$ is visited more often:
\begin{equation}
M(v)
=
\max
\left\{
1,
k N(v)^{\alpha}
\right\}.
\end{equation}
The coefficients are set as $k=1.0$, $\alpha=0.5$ and $C=3$. Given the selected parent node $v$, \textsc{ProposeSuccessor} returns a random untested successor mode $c^+$, defining the candidate sequence
\begin{equation}
\mathcal C^+
=
\mathcal C(v) \oplus c^+ .
\end{equation}

Before admission to the tree, \textsc{Verify} applies the selected feasibility filters $\mathcal F$. We consider the following filters, ordered from cheaper to more expensive:
\begin{enumerate}
\item contact-mode feasibility, $\text{M}(c^+)$, using \eqref{eq:graph_pruning},
\item transition-edge feasibility, $\text{E}(c(v),c^+)$, using \eqref{eq:graph_pruning},
\item kinematic sequence optimization, $\text{KSO}(\mathcal C^+)$, using \eqref{eq:seqik},
\item trajectory optimization, $\text{TO}(\mathcal C^+)$, using \eqref{eq:TO}.
\end{enumerate}
The mode filter $\text{M}$ evaluates the proposed successor mode $c^+$, while the edge filter $\text{E}$ evaluates the transition from the terminal mode $c(v)$ of the parent sequence to $c^+$. The sequence-level filters $\text{KSO}$ and $\text{TO}$ evaluate the complete candidate sequence $\mathcal C^+$.

The four search variants evaluated in this work use
\begin{equation}
\begin{aligned}
\mathcal F_1 &= \bigl(\text{KSO}\bigr),
&
\mathcal F_2 &= \bigl(\text{M},\text{E},\text{KSO}\bigr),
\\
\mathcal F_3 &= \bigl(\text{M},\text{E},\text{KSO},\text{TO}\bigr),
&
\mathcal F_4 &= \bigl(\text{TO}\bigr).
\end{aligned}
\end{equation}
Within each variant, the filters are applied from left to right and verification terminates at the first failed filter. Reusable outcomes are stored in the feasible and infeasible caches, $\mathcal K_{\mathrm{feas}}$ and $\mathcal K_{\mathrm{infeas}}$, so that previously evaluated checks do not need to be solved again. If all active filters succeed, the candidate is added to the tree and assigned the cost $J$ returned by the final filter in $\mathcal F$. Successful goal-reaching sequences are added to $\mathcal S_{\mathrm{goal}}$, and the search continues until the time budget $B$ is exhausted.

\begin{algorithm}[t]
\caption{Feasibility-Guided Contact-Mode Tree Search}
\label{alg:feasible_tree_search}
\begin{algorithmic}[1]
\Require initial mode $c_0$, time budget $B$, feasibility filters $\mathcal F$
\State $\mathcal T \gets \textsc{InitializeTree}(c_0)$
\State $\mathcal S_{\mathrm{goal}} \gets \varnothing$
\State $\mathcal K_{\mathrm{feas}},\mathcal K_{\mathrm{infeas}} \gets \varnothing,\varnothing$

\While{$\textsc{ElapsedTime}() < B$}
\State $v \gets \textsc{Select}(\mathcal T)$
\Comment{cost-based UCT with progressive widening}

\State $c^+ \gets \textsc{ProposeSuccessor}(v)$
\State $\mathcal C^+ \gets \mathcal C(v) \oplus c^+$

\State $(\mathrm{feasible},J)
\gets
\textsc{Verify}
(\mathcal C^+,\mathcal F,
\mathcal K_{\mathrm{feas}},
\mathcal K_{\mathrm{infeas}})$

\If{$\neg\mathrm{feasible}$}
    \State \textbf{continue}
\EndIf

\State $\textsc{AddChild}(\mathcal T,v,c^+,J)$

\If{$\textsc{Goal}(\mathcal C^+)$}
    \If{$\text{TO}\in\mathcal F$}
        \State $\mathrm{goalFeasible}\gets\mathrm{true}$
    \Else
        \State $\mathrm{goalFeasible}
        \gets \textsc{TrajectoryOpt.}(\mathcal C^+)$
    \EndIf

    \If{$\mathrm{goalFeasible}$}
        \State $\mathcal S_{\mathrm{goal}}
        \gets
        \mathcal S_{\mathrm{goal}}
        \cup
        \{\mathcal C^+\}$
    \EndIf
\EndIf

\EndWhile

\State \textbf{return } $\mathcal S_{\mathrm{goal}}$
\end{algorithmic}
\end{algorithm}

%===============================================================================
\section{Evaluation}
\label{sec:evaluation}

We evaluate our method in three settings. First, using human-specified contact sequences, we compare the computational cost and constraint coverage of the KSO and TO formulations and demonstrate real-world execution of the resulting trajectories on a humanoid robot. Second, we evaluate the proposed feasibility-guided tree search in Algorithm~\ref{alg:feasible_tree_search}, measuring its exploration performance and solution discovery under a fixed time budget. Finally, we evaluate KSO as a feasibility filter for contact plans generated by an LLM across several scene exploration tasks.

\subsection{Qualitative analysis on human-defined contact sequences}

To evaluate the proposed optimization hierarchy on human-defined contact plans, we design eight contact-mode sequences with varying lengths and levels of complexity. Each mode in the TO is discretized using 20 optimization knots. For each sequence, we report the TO and KSO solve times, the reduction in decision variables from TO to KSO, and the fraction of TO constraint types represented in the KSO. The resulting data is presented in Table \ref{tab:qualitative_analysis}. The sequences marked with an asterisk are additionally executed on the real robot using an RL-based trajectory-tracking controller~\cite{dhedin2026dynaretargetdynamicallyfeasibleretargetingusing} (experiments are captured in the accompanying video). Snapshots of the trajectory and real world execution of the Double Catch task can be seen in Fig.~\ref{fig:teaser}.

Across the eight sequences, the KSO is approximately two orders of magnitude faster than the TO while retaining, on average, 70.0\% of the constraint types present in the TO. This combination of high constraint coverage and an average 74.8$\times$ reduction in decision variables makes the KSO a fast yet expressive feasibility check. The constraints omitted by the KSO are those associated with dynamics, since it optimizes one configuration per mode transition rather than a fully discretized trajectory.

\subsection{Feasibility-guided tree search}
\begin{figure}
    \vspace*{2pt}
    \centering
    \includegraphics[width=0.4\textwidth]{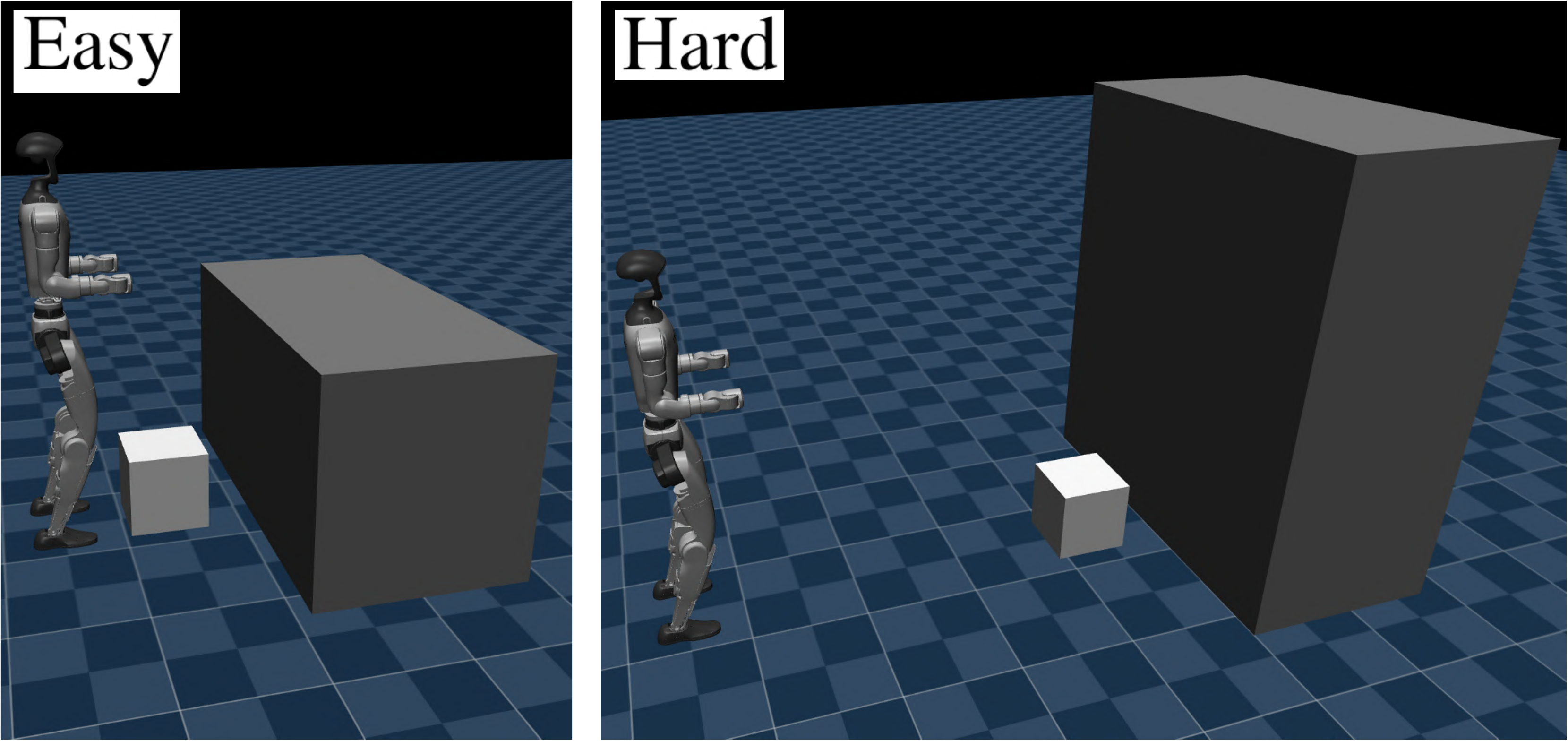}
    % \vspace{-4mm}
    \caption{\textbf{Tree Search Evaluation Scenes.} The feasibility-guided tree search is evaluated on two box-placement tasks with the same contact-mode search space but different levels of kinematic and dynamic difficulty. In both tasks, the objective is to place the white box on the platform.}
    \label{fig:mcts_evals}
\end{figure}

We evaluate the proposed feasibility-guided tree search Algorithm~\ref{alg:feasible_tree_search} on two variants of the box-placement task shown in Figure~\ref{fig:mcts_evals}. Each variant is evaluated over five runs with different seeds controlling the random successor selection, with a two-hour time budget per run and a max depth of 5 switches. As a baseline, we use tree search with TO as the only feasibility check, which is similar to the exploration strategies used in~\cite{toussaint2018differentiable,sleiman2023versatile,ciebielski2025task}. Table~\ref{tab:mcts_seqik_filter} reports the search statistics for the different filter sets~$\mathcal F$. The allowed contact interfaces are listed in Table~\ref{tab:allowed_contacts}, yielding a maximum branching factor of 108.

Across both task variants, searches using the proposed feasibility filters outperform the TO-only baseline in solution discovery and tree expansion. Filters that exclude TO during node expansion explore substantially faster and invoke TO only after reaching the discrete goal, indicating that KSO can serve as an effective proxy for guiding search. This advantage is especially pronounced in the hard task where the TO-only baseline finds no solutions and expands only 12.8 nodes on average, whereas the $\text{M},\text{E},\text{KSO}$ variant expands 814.6 nodes and finds 26.4 solutions on average.

The hard task also highlights the benefit of mode and edge filtering in addition to KSO. Since contact modes and transitions are repeatedly revisited during search, cached infeasibility results allow previously checked queries to be rejected without resolving the corresponding optimization problems. This substantially reduces redundant exploration and explains the large performance gap between the $\text{KSO}$ and $\text{M},\text{E},\text{KSO}$ variants.

\begin{table*}[t]
\centering
\caption{Computational performance and constraint coverage on human-specified contact sequences.}
\label{tab:qualitative_analysis}
\resizebox{\linewidth}{!}{%
\begin{tabular}{lcccccccc@{\hspace{1.2em}}c}
\toprule
Metric/Task 
& Climb 1 
& Climb 2 
& Pick Place 1\textsuperscript{*}
& Pick Place 2\textsuperscript{*}
& Toss to Table\textsuperscript{*}
& Double Catch\textsuperscript{*}
& Triple Catch
& Juggle\textsuperscript{*}
& \textbf{Avg.} \\
\midrule
Num. Modes $K$
& 6 & 11 & 6 & 11 & 4 & 8 & 11 & 10
& \textbf{8.4} \\
TO time [s]
& 50.59 & 106.01 & 21.36 & 83.40 & 9.73 & 91.53 & 117.62 & 37.33
& \textbf{64.70} \\
KSO time [s]
& 0.34 & 1.93 & 0.37 & 0.90 & 0.14 & 0.31 & 0.42 & 0.43
& \textbf{0.60} \\
% Speedup [$\times$]
% & 151.0 & 54.9 & 58.1 & 92.7 & 69.5 & 297.2 & 279.4 & 86.4
% & \textbf{136.1} \\
Dec.-var. reduction [$\times$]
& 74.9 & 74.4 & 74.9 & 74.9 & 74.9 & 74.9 & 74.9 & 74.9
& \textbf{74.8} \\
Constraint coverage [\%]
& 70.2 & 70.6 & 70.2 & 71.3 & 70.2 & 69.1 & 69.1 & 69.1
& \textbf{70.0} \\
\bottomrule
\end{tabular}
 }
\end{table*}

\begin{table*}[t]
\centering
\caption{Feasibility-guided tree search performance across filter configurations.}
\label{tab:mcts_seqik_filter}
\resizebox{\linewidth}{!}{%
\begin{tabular}{l*{8}{c}}
\toprule
& \multicolumn{4}{c}{Easy}
& \multicolumn{4}{c}{Hard} \\
\cmidrule(lr){2-5}
\cmidrule(lr){6-9}
Metric/Task
& KSO & M,E,KSO & M,E,KSO,TO & TO
& KSO & M,E,KSO & M,E,KSO,TO & TO \\
\midrule

No. sol.
& $73.6 \pm 19.7$ & $83.8 \pm 6.5$ & $8.0 \pm 3.7$ & $20.0 \pm 5.0$
& $0.4 \pm 0.9$ & $26.4 \pm 9.7$ & $1.2 \pm 0.8$ & $0.0 \pm 0.0$ \\

Time to first sol. [s]
& $192 \pm 96$ & $194 \pm 74$ & $1455 \pm 989$ & $951 \pm 583$
& $2671 \pm \mathrm{n/a}$ & $2763 \pm 1549$ & $5399 \pm 868$ & n/a \\

KSO attempts
& $410.6 \pm 27.1$ & $410.0 \pm 18.7$ & $230.8 \pm 16.4$ & $0.0 \pm 0.0$
& $1660.8 \pm 55.4$ & $1415.0 \pm 77.4$ & $437.4 \pm 13.0$ & $0.0 \pm 0.0$ \\

TO attempts
& $182.0 \pm 4.2$ & $176.2 \pm 7.4$ & $223.0 \pm 13.6$ & $208.4 \pm 8.3$
& $5.4 \pm 2.4$ & $79.6 \pm 7.3$ & $164.0 \pm 13.8$ & $216.2 \pm 5.5$ \\

Total tree nodes
& $393.8 \pm 24.3$ & $380.8 \pm 22.3$ & $90.6 \pm 5.5$ & $87.8 \pm 12.2$
& $218.2 \pm 24.0$ & $814.6 \pm 78.9$ & $52.8 \pm 7.7$ & $12.8 \pm 2.2$ \\

\bottomrule
\end{tabular}
}
\end{table*}

\begin{table}[t]
\centering
\caption{LLM Exploration}
\label{tab:llm_exploration}
\resizebox{\linewidth}{!}{%
\begin{tabular}{lcccc}
\toprule
Metric/Task & 1-a & 1-b & 2-a & 2-b\\
\midrule
KSO-feas. & $30.8 \pm 11.1$ & $4.6 \pm 6.4$ & $47.0 \pm 9.9$ & $20.6 \pm 28.0$ \\
TO-feas. & $8.8 \pm 5.0$ & $0.4 \pm 0.9$ & $10.2 \pm 8.0$ & $2.8 \pm 5.7$ \\
FNR [\%] $\downarrow$ & $0.0 \pm 0.0$ & $0.0 \pm 0.0$ & $1.3 \pm 2.6$ & $0.0 \pm 0.0$ \\
FPR [\%] $\downarrow$ & $24.2 \pm 10.4$ & $4.2 \pm 5.8$ & $41.4 \pm 8.3$ & $19.5 \pm 26.1$\\
Speedup [$\times$] $\uparrow$ & $3.8 \pm 1.4$ & $15.5 \pm 10.2$ & $2.4 \pm 0.5$ & $7.4 \pm 4.3$ \\
\bottomrule
\end{tabular}
}
\end{table}

\begin{table}
\centering
\caption{Allowed contact interactions, relative to initial scene state.}
\label{tab:allowed_contacts}
\begin{tabular}{ll}
\toprule
Interface & Allowed contacts \\
\midrule
Left hand & box left, box front, free \\
Right hand & box right, box rear, free \\
Left foot & floor, free \\
Right foot & floor, free \\
Box bottom & floor, tabletop, free \\
\bottomrule
\end{tabular}
\end{table}

\begin{figure}
    \centering
    \includegraphics[width=0.48\textwidth]{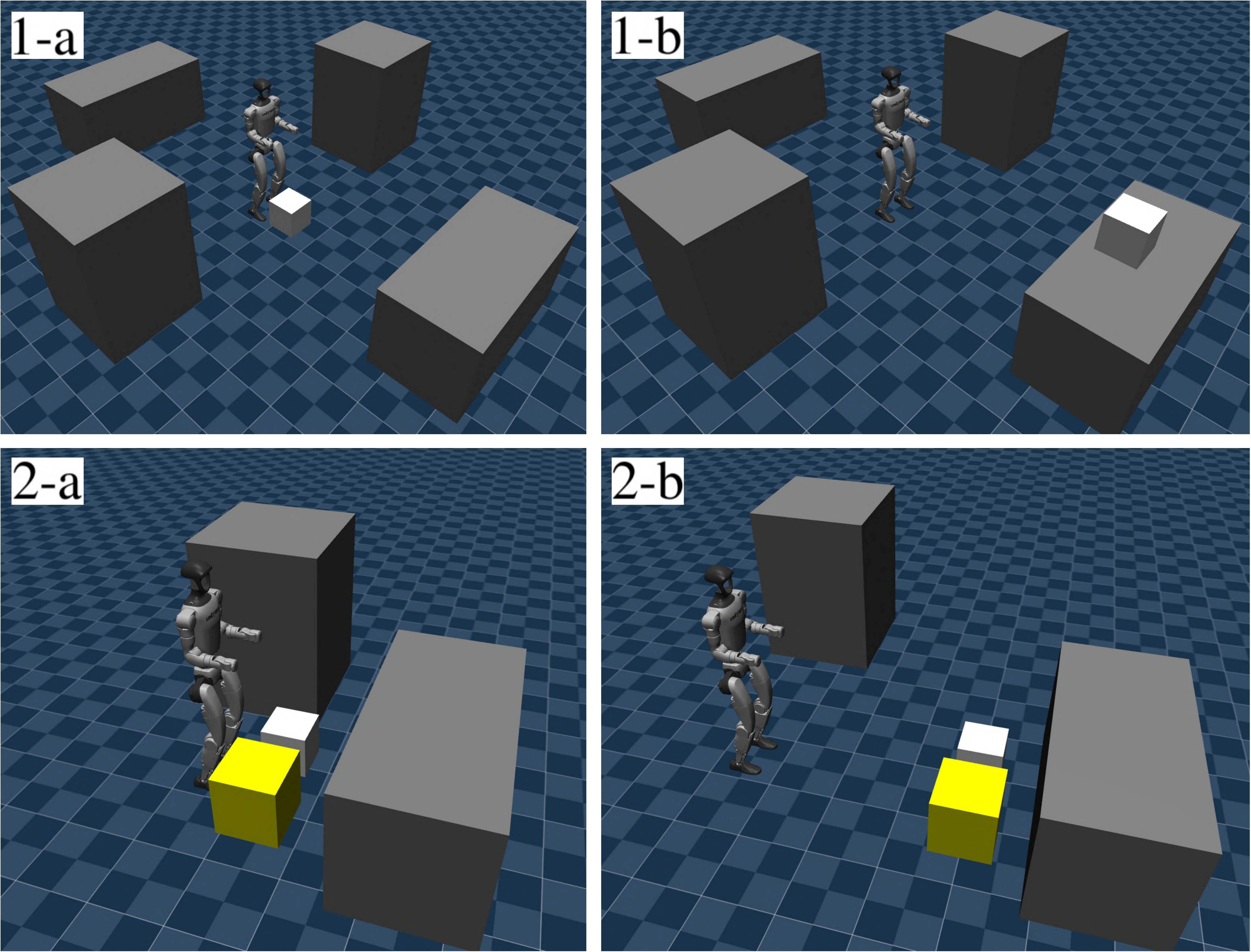}
    % \vspace{-4mm}
    \caption{\textbf{LLM Contact-Plan Sampling Scenes.} The KSO feasibility filter is evaluated on LLM-generated contact plans across four task scenarios. For each task, the \textit{-b} variant is more challenging than the corresponding \textit{-a} variant. The LLM is prompted to output contact plans that explore the scene in qualitatively different ways.}
    \label{fig:llm_evals}
\end{figure}

\subsection{LLM-based Contact Plan Sampling}
We evaluate the KSO as a feasibility filter for contact plans generated by \textit{GPT-5.5}~\cite{openai2026chatgpt55}. Given a scene and goal description, the LLM is prompted to generate 100 diverse contact plans. We repeat this stochastic generation five times with the same prompt for each scene in Fig.~\ref{fig:llm_evals} and report the results in Table~\ref{tab:llm_exploration}. For each plan, we solve both the KSO and TO and evaluate the KSO as a classifier of TO feasibility using its false-negative and false-positive rates. We also report the computational speedup obtained when the KSO is used as a feasibility filter before TO. Each scene has an easier \textit{-a} variant and a more challenging \textit{-b} variant.

KSO filtering provides a speedup in all tasks, with larger gains in the harder variants. At the same time, the false-negative rate remains near zero, indicating that KSO rarely rejects plans that are feasible under TO. This is consistent with the shared constraint structure of the two formulations where the KSO preserves most of the geometric and contact constraints enforced by TO, while omitting the additional dynamic constraints and trajectory variables. The few observed false negatives were primarily associated with poor initialization or insufficient solver iterations in the KSO.

%===============================================================================

\section{Conclusion}
\label{sec:conclusion}
In this work, we presented \textit{FARO}, a feasibility-aware framework for rapidly discovering dynamic humanoid loco-manipulation behaviors from candidate contact-mode sequences. \textit{FARO} hierarchically filters candidates using contact mode, transition, kinematic sequence, and full trajectory optimization checks. Early pruning and cached feasibility results reduce expensive trajectory optimizations while preserving promising contact sequences.

Experiments across feasibility-guided tree search, LLM-generated, and human-designed plans show that the mode/edge and kinematic sequence optimizations capture most relevant constraints at substantially lower cost than full trajectory optimization, enabling faster and broader exploration of candidate contact plans and improving solution discovery in challenging scenarios while producing very few false-negative classifications.

%===============================================================================

These results establish the practical value of our feasibility-aware motion optimization, while also highlighting several directions for further development. The current formulation assumes planar contact interfaces and could be extended to richer geometries using differentiable signed-distance representations such as~\cite{dietz2025smoothed}. The optimization hierarchy is designed around increasingly restrictive feasibility conditions, and we expect the corresponding feasible sets to be nested, although a formal proof of this property is left to future work. Finally, scaling to substantially longer-horizon tasks will likely require integrating \textit{FARO} with higher-level task and motion planning or program search~\cite{taouil2026motiondisco} to manage the combinatorial complexity of contact-space search.

%===============================================================================

\bibliographystyle{IEEEtran}
\bibliography{example}

% Biography text here without a photo.

\end{document}